\documentclass[twoside,journey]{IEEEtran}
\usepackage{makecell}
\usepackage{hyperref}
\usepackage{array}
\usepackage{graphicx,amssymb,amsmath}
\usepackage{multicol}
\usepackage[noadjust]{cite}
\usepackage{setspace}
\usepackage{subcaption}
\usepackage{graphicx}
\usepackage{float}
\usepackage {url}
\usepackage{stfloats}
\usepackage{amsthm,pifont}
\usepackage{flushend}
\usepackage{cases,subeqnarray}
\usepackage{bm,multirow,bigstrut}
\usepackage{amsmath, amsthm, amssymb}
\usepackage{textcomp}
\usepackage{latexsym,bm}
\usepackage{booktabs}
\usepackage{xcolor}
\usepackage{mathtools}
\usepackage{dsfont}
\usepackage{extarrows}
\usepackage{epsfig}
\usepackage{epsfig}
\usepackage{epstopdf}
\usepackage[noend]{algpseudocode}
\usepackage{algorithmicx,algorithm}
\usepackage{makecell}
\usepackage{colortbl}
\usepackage{booktabs}
\usepackage{graphicx} 
\usepackage{array}    
\usepackage{hyperref} 
\usepackage{booktabs}
\usepackage{enumitem}
\usepackage{diagbox}
\usepackage{tikz}
\usetikzlibrary{trees,shadows.blur}
\usepackage{forest}
\theoremstyle{plain}

\theoremstyle{plain}

\usepackage{amsmath}

\newcommand{\ignore}[1]{{{\color{yellow} }}}
\definecolor{blue-green}{rgb}{0.0, 0.87, 0.87}
\IEEEoverridecommandlockouts
\begin{document}

\title{World Models for Cognitive Agents: Transforming Edge Intelligence in Future Networks}
\author{Changyuan Zhao, Ruichen Zhang, Jiacheng Wang, Gaosheng Zhao, Dusit Niyato~\IEEEmembership{Fellow,~IEEE}, \\Geng Sun,  
Shiwen Mao~\IEEEmembership{Fellow,~IEEE},  
Dong In Kim,~\IEEEmembership{Life Fellow,~IEEE}
\thanks{C. Zhao is with the College of Computing and Data Science, Nanyang Technological University, Singapore, and CNRS@CREATE, 1 Create Way, 08-01 Create Tower, Singapore 138602 (e-mail: zhao0441@e.ntu.edu.sg).}
    \thanks{R.~Zhang, J.~Wang, and D. Niyato are with the College of Computing and Data Science, Nanyang Technological University, Singapore (e-mail: ruichen.zhang@ntu.edu.sg, jiacheng.wang@ntu.edu.sg, dniyato@ntu.edu.sg).}
    
        \thanks{G. Sun is with College of Computer Science and Technology, Jilin University, China 130012, (e-mail: sungeng@jlu.edu.cn).}
        \thanks{S. Mao is with the Department of Electrical and Computer Engineering,
Auburn University, Auburn, USA (e-mail: smao@ieee.org).}
    \thanks{G. Zhao and D. I. Kim are with the Department of Electrical and Computer Engineering, Sungkyunkwan University, Suwon 16419, South Korea (e-mail: gaosheng@skku.edu, dongin@skku.edu).}}

\maketitle
\vspace{-1cm}

\begin{abstract}
World models are emerging as a transformative paradigm in artificial intelligence, enabling agents to construct internal representations of their environments for predictive reasoning, planning, and decision-making. By learning latent dynamics, world models provide a sample-efficient framework that is especially valuable in data-constrained or safety-critical scenarios. In this paper, we present a comprehensive overview of world models, highlighting their architecture, training paradigms, and applications across prediction, generation, planning, and causal reasoning. We compare and distinguish world models from related concepts such as digital twins, the metaverse, and foundation models, clarifying their unique role as embedded cognitive engines for autonomous agents. We further propose Wireless Dreamer, a novel world model-based reinforcement learning framework tailored for wireless edge intelligence optimization, particularly in low-altitude wireless networks (LAWNs). Through a weather-aware UAV trajectory planning case study, we demonstrate the effectiveness of our framework in improving learning efficiency and decision quality. 
\end{abstract}
\begin{IEEEkeywords}
Edge intelligence, wireless communications, low-altitude wireless networks
\end{IEEEkeywords}
\IEEEpeerreviewmaketitle

\section{Introduction}\label{intro}


In the science fiction film \textit{The Matrix}, the eponymous system serves as a platform that forecasts future outcomes based on the protagonist’s decisions. This fictional technology presents a compelling vision: if artificial intelligence (AI) models can accurately internalize the workings of the real world, they could unlock the ability to perform complex tasks through predictive reasoning.
This idea is now gaining significant attention in AI research: \textit{how can AI, like humans, learn the fundamental laws governing our world and make informed predictions?}


Motivated by this idea, \textit{world models} have emerged as a class of AI systems that learn an internal model of the environment’s dynamics, such as its physical laws and spatial properties\footnote{https://www.nvidia.com/en-sg/glossary/world-models}. By predicting future observations and rewards, they provide an agent with an ``imagination” space for various tasks, including planning and control.
Concretely, world models are typically built upon generative AI (GenAI) frameworks trained to reconstruct past observations and forecast future trajectories.
By implicitly capturing the rules and patterns of the environment, the world model estimates what is likely to happen next. Rather than relying solely on immediate sensor data, the agent uses this internal model as a forward-looking guide, much like a chess player thinking several moves ahead or a driver intuitively expecting a pedestrian to jaywalk. 
In essence, the world model grants the AI agent a cognitive capacity to perceive, anticipate, and reason about its environment, enabling it to make more informed and adaptive decisions under uncertainty.

In contrast with traditional AI models built for single or a few tasks, a world model offers a general framework whose main advantage is its ability to generalize to diverse downstream tasks, including generation, planning, and reasoning. While the world model acts as the brain that provides the agent with an understanding of the real world, its specific focus varies depending on the task.
For generation tasks, world models act as imagination engines, creating plausible future scenarios and generating remarkably realistic video sequences that adhere to physical laws, emphasizing that these models capture fundamental principles~\cite{zhu2024sora}.
In terms of planning, such as autonomous driving, a world model allows an agent to mentally simulate action sequences and predict unexpected events, enhancing its robustness in real-world execution~\cite {guan2024world}.
When it comes to reasoning, the agent leverages world models to anticipate the consequences of its decisions and infer hidden causes, strengthening its decision-making with almost human-like foresight
\cite{ding2024understanding}.



Given these strengths, world models are increasingly recognized as foundational for advanced cognitive agents. As humans use mental models to understand and navigate their environment, AI systems embed world models at the core of their cognitive architecture to guide both perception and action. Many state-of-the-art systems, such as Wayve’s GAIA model for safer assisted and autonomous driving\footnote{https://wayve.ai/thinking/gaia-2/}, exemplify this approach.
Looking ahead to next-generation networks, where countless smart devices and autonomous systems operate at the edge, world models can be transformative. Edge agents, from self-driving cars and delivery drones to intelligent sensors, must often make instant decisions locally, without relying on constant cloud connectivity. By equipping them with robust world models, these agents can predict, plan, and act directly on-site, improving responsiveness and reducing dependence on cloud back-ends. Companies such as NIO have already embraced this paradigm, developing the NIO World Model specifically for on-car AI predictions\footnote{https://www.nio.cn/smart-technology/20241120002}.

Building on these foundations, this paper provides a comprehensive perspective on world models, covering from concept to deployment. We then introduce several applications and highlight their potential impact in the realm of edge intelligence.
Furthermore, we propose a world model framework, Wireless Dreamer, for wireless edge intelligence optimizations. Unlike traditional optimization or reinforcement learning (RL) methods, our approach leverages a world model to recover and predict data for more effective decision-making in low-altitude wireless networks (LAWNs). \textit{To the best of our knowledge}, this is the first work to explore the integration of world models for wireless edge intelligence and decision-making in LAWNs.
The key contributions of this work are summarized as follows:
\begin{itemize}
    \item We delve into the concept and deployment of world models, highlighting their advantages. Subsequently, we provide a detailed comparison between world models and other approaches, emphasizing their uniqueness.
    
    \item We propose a world model framework, Wireless Dreamer, for wireless edge intelligence optimizations, which leverages world models to enhance and predict perception data, supporting effective decision-making.
    
    \item We consider a weather-aware UAV-assisted wireless communication task in LAWNs through UAV trajectory planning, as a case study.
    We employ the Wireless Dreamer framework to solve the problem and demonstrate the effectiveness of this framework by benchmarking it with several baselines.
\end{itemize}

\section{Overview of World Models}


\begin{figure*}[htp]
    \centering
    \includegraphics[width= 0.87\linewidth]{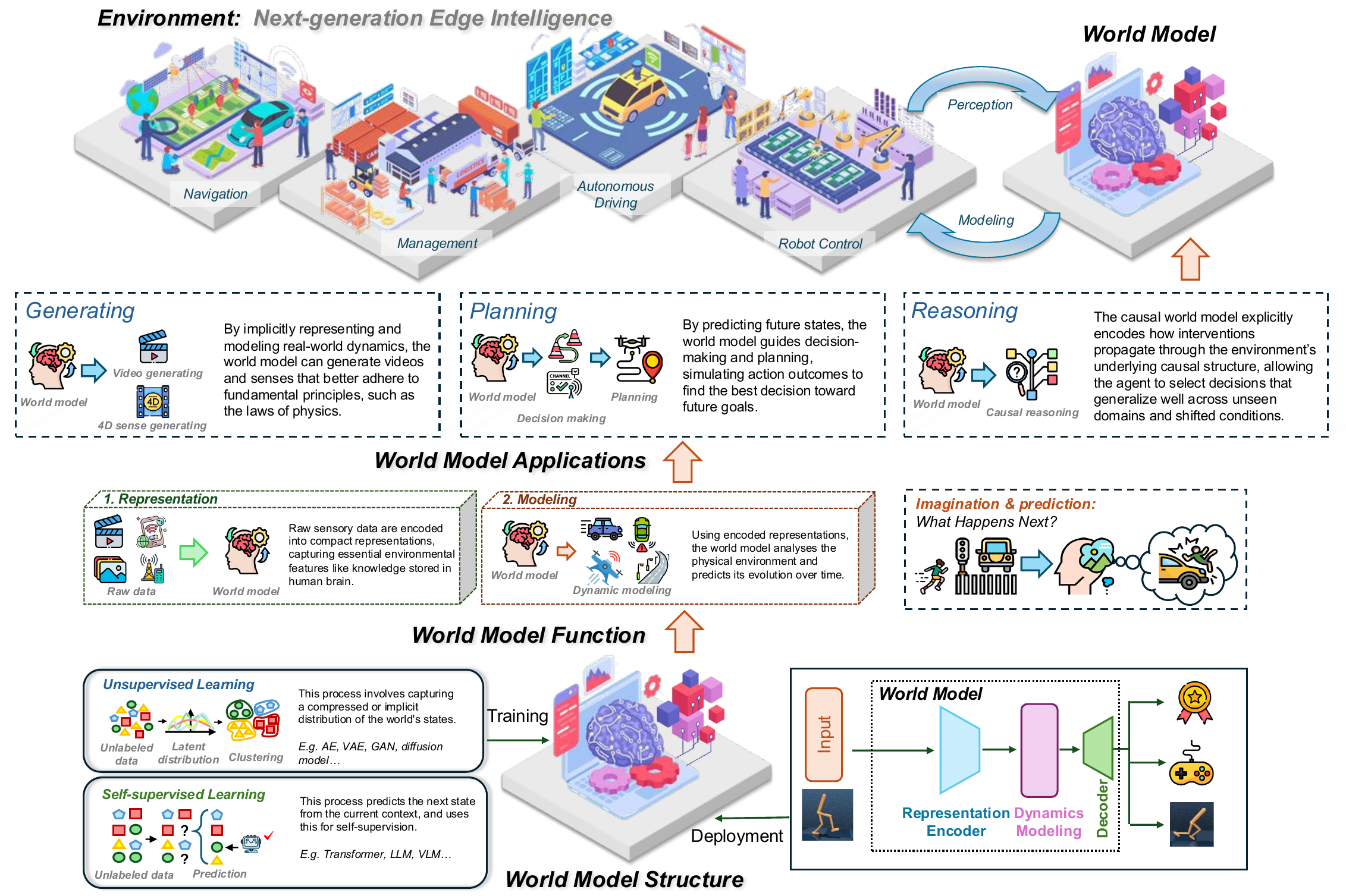}
    \caption{Illustration of world models for next-generation edge intelligence. 
    The bottom part shows the workflow and structure of world models, including their training and deployment.
    The middle part illustrates the core functions of world models.
    The top part highlights the three core applications of world models and their potential application in a real-world environment.
    }
    \label{fig:world}
\end{figure*}

\subsection{General Concept of World Models}

A world model is an internal representation that an intelligent system uses to understand and predict environmental dynamics. It serves as a cognitive map or simulator, encoding the external world's state in compressed latent variables and modeling its temporal evolution, allowing the agent to anticipate outcomes and plan actions accordingly \cite{ding2024understanding}.
World models can be broadly classified into two functional categories:
\begin{itemize}
    \item \textbf{Implicit Representation:}
    These models capture environmental dynamics as latent representations, supporting more informed decision-making through an LLMs, the transformation of real-world phenomena into implicit knowledge enhances decision-making by enabling the models to represent and reason about complex world knowledge.
     
    \item \textbf{Dynamics Modeling:}
    These models use current observations to predict future events by modeling continuous spatial and temporal dynamics. For instance, large vision models (LVMs) and vision-language models (VLMs) \cite{zhang2024vision}, such as Sora and Kling AI, can generate video content that reflects physical laws, effectively modeling dynamic processes including motion trajectories, obstacle interactions, and other real-world physical behaviors.
\end{itemize}

Based on these functional categories, the world model integrates knowledge learning and environment modeling to create a comprehensive system capable of complex interactions with dynamic environments, supporting various downstream tasks.
As depicted in Fig. \ref{fig:world}, the general workflow and structure of world models includes:
\begin{itemize}
    \item \textbf{Encoder:}
    First, the raw sensory inputs
    are processed and encoded into a structured internal representation. Through dimensionality reduction and feature extraction, these diverse and complex observations are transformed into compact and meaningful representations that reflect the essential characteristics of the environment.
    

    \item \textbf{Modeling:}
    Utilizing encoded representations, the world model effectively models and analyzes the physical environment. Subsequently, the model predicts the evolution of the physical environment over time.
    Accurate dynamics modeling allows for imagining future states, enabling proactive and anticipatory responses from agents.
    
    \item \textbf{Decoder:}
    By leveraging predictive understanding of future states, the model ultimately supports diverse downstream tasks such as autonomous driving and robot control, through decoding its implicit representations.
    Incorporating techniques such as RL can further enhance this process.

\end{itemize}


It is worth noting that the last part, the decoder, is a downstream task-dependent module, so the same world model can be adapted to different downstream tasks by swapping or fine‑tuning different decoders \cite{hafner2020mastering}.



\subsection{Deployment of World Models}

Training and deploying world models typically leverage GenAI approaches, where the model learns implicit representations of the input data and generates content that mimics the same underlying data distribution.
When the input consists of real-world environmental data, such as radar, LiDAR, or videos, the GenAI model can generate new content to predict various aspects of the environment based on a given prompt that specifies which elements are most important.
Two predominant training paradigms are self-supervised learning and unsupervised learning, both of which exploit unlabeled data to learn the model, as shown in Fig. \ref{fig:world}. 
\begin{itemize}
    \item \textbf{Unsupervised Learning:}
    The world models learn representation in the environment data without explicit prediction tasks.
    Common approaches include autoencoders (AEs), variational autoencoders (VAEs), and diffusion models, which learn abstract state representations that serve as a foundation for further analysis \cite{zhao2024generative}. For example, in semantic communication, an agent might compress high-dimensional camera images into a latent state that retains essential details about the scene. These learned representations enable more efficient processing, transmission, and decision-making in complex environments.

    \item \textbf{Self-supervised Learning:}
    The world models learn surrogate tasks from raw data, such as predicting the next state from the current context, and use this for self-supervision. Once trained, they can generate future trajectories by iteratively predicting each step \cite{zhao2025generative}.
    Recently, Transformer-based sequence models, including LLMs, have been applied, treating sequences of states as ``sentences," predicting one token at a time. For example, a world model could forecast traffic load and channel conditions at base stations based on historical data, enabling intelligent spectrum allocation and power control.
\end{itemize}

The key distinction between these methods is that unsupervised learning captures the structure or appearance of the world, while supervised learning focuses on predicting the next event. These align with the two roles of a world model: \textbf{Implicit Representation} and \textbf{Prediction}.

\begin{table*}[ht]
    \centering
    \caption{Comparison of World Models with Other Models and Their Downstream Applications}
    \label{tab:pdf-table}
    \includegraphics[width= 0.90\linewidth]{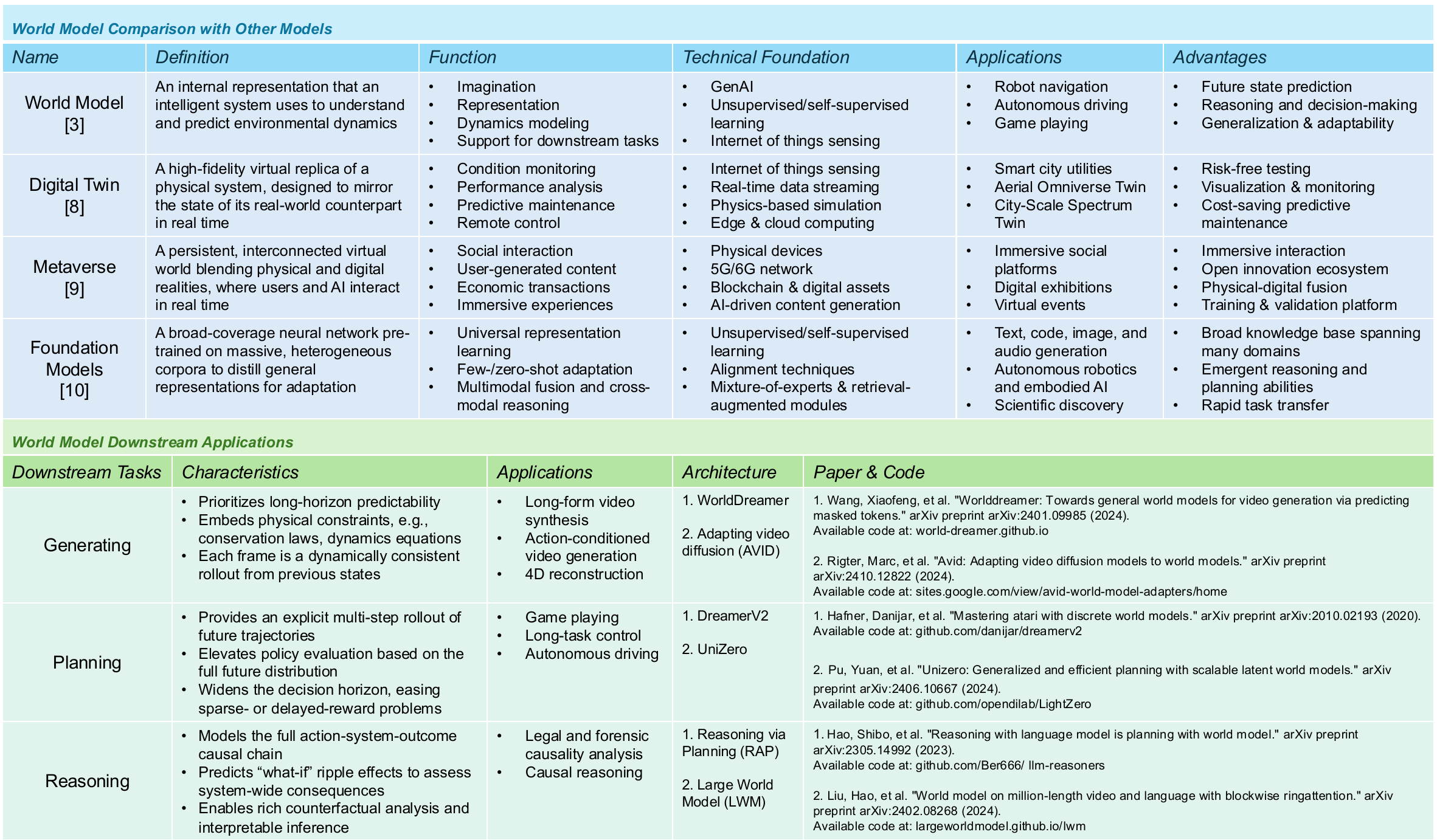} 
\end{table*}

\subsection{Comparison with Other Models}


\subsubsection{World Models vs. Digital Twins}

A digital twin is a high-fidelity virtual replica of a physical system, designed to mirror the state of its real-world counterpart in real time~\cite{khan2022digital}. It serves as an external, synchronized model used by humans or AI for testing and issue detection. 
In contrast, a world model is an internal, abstract representation learned by AI agents, focused on achieving agents' goals. 
While a digital twin aims for detailed accuracy, often as a stand-alone system, a world model prioritizes efficient decision-making and adaptability, integrated within the agent’s reasoning.

\subsubsection{World Models vs. the Metaverse}

The metaverse refers to a persistent, interconnected virtual world blending physical and digital realities, where users and AI interact in real time~\cite{xu2022full}. It hosts both digital twins of real-world objects and fictional elements, aiming for immersive experiences rather than strict realism. 
While the metaverse is an external, shared space used for interaction and experimentation, world models are internal tools for decision-making. Therefore, the metaverse can serve as a platform to build and test world models, offering a rich, high-fidelity environment for training network control systems.

\subsubsection{World Models vs. Foundation Models}

The foundation model is a broad-coverage neural network pre-trained on massive, heterogeneous corpora, including text, images, and code, to distill general representations ready for zero-/few-shot adaptation \cite{bommasani2021opportunities}. In contrast, a world model is embodied within an agent, compressing a specific environment into latent states and transition dynamics to guide planning and control. 
Foundation models provide general knowledge that can help world models understand perception or language, while the policy execution gained through world model actions can improve foundation models by enhancing their reasoning and real-world alignment in turn.



To summarize, we can posit an analogy related to a city-building project. The world model represents the engineer's mental sandbox for simulating various outcomes. The digital twin reflects the real-time construction site. The metaverse offers a shared virtual space for collaboration among team members. Lastly, the foundation model serves as a knowledgeable advisor, providing guidance to inform decision-making based on broad expertise.
For a more intuitive understanding, we present a comprehensive comparison in the Table \ref{tab:pdf-table}.

\section{Applications of World Models}




\subsection{Predicting}



The core application of the world model is to learn implicit expression and prediction.
Formally, a world model is trained to approximate the latent Markov process,
thereby internalizing how the environment evolves from one instant to the next.
For example, in the seminal work \cite{ha2018world}, a recurrent network was used as a world model to predict the next state and reward based on the agent’s action in the CarRacing game. 
Through ablation studies, the authors demonstrated that removing this world model led to a 30\% drop in gaming score and a significant increase in variance \cite{ha2018world}. 
Unlike traditional RL, which directly maps observations to actions, a world model-based approach leverages predictive modeling to anticipate outcomes. These predictive capabilities not only improve action selection but also support downstream tasks, such as control, forecasting, and anomaly detection, which are further explored throughout this paper.

\subsection{Generating}

For generative tasks of world models, the central objective is to produce coherent temporal data, video streams, and 4‑D scene reconstructions. 
Unlike conventional video generation networks, a world model prioritizes long‑horizon predictability and explicit adherence to underlying physical laws.
For example, WorldDreamer is a video world model that learns universal motion and physical dynamics, which are embedded in visual signals \cite{wang2024worlddreamer}.
Built on a Spatial–Temporal Patchwise Transformer, the model restricts attention to localized windows in both space and time, allowing it to capture fine‑grained visual dynamics efficiently.
Consequently, WorldDreamer supports several generating tasks, including video inpainting, video stylization, and action‑to‑video generation, within a single unified framework. Remarkably, it can synthesize a high-quality 24‑frame video at $192\times320$ resolution in just 3 seconds on a single A800 GPU \cite{wang2024worlddreamer}.

\subsection{Planning}

For planning tasks, the chief objective is to synthesize anticipatory action sequences that maximize long‑term returns.
Unlike conventional Markov‑decision processes (MDPs) or RL, where each policy update relies solely on the current state, a world model furnishes the planner with a prediction of future contingencies, effectively widening the decision horizon beyond one‑step transition probabilities.
For instance, 
DreamerV2 learns long‑horizon behaviour entirely within this learned world model, employing an actor–critic framework to learn the policy~\cite{hafner2020mastering}. The actor selects actions by imagining sequences of compact latent states, while the critic aggregates the predicted future rewards to capture returns that lie beyond the planning horizon.
Notably, DreamerV2 attains human‑level performance on the Atari benchmark, achieving a gamer‑normalized median score above 2.0 compared with the human gamer baseline of 1.0
\cite{hafner2020mastering}.




\subsection{Reasoning}

For reasoning tasks, the primary goal is to reveal the underlying causal mechanisms driving the environment’s behavior. Unlike conventional statistical methods that merely capture correlations between surface-level features, a world model simulates and anticipates how actions will ripple through the system.
For example, the authors in \cite{hao2023reasoning} proposed a LLM reasoning framework called Reasoning via Planning (RAP). RAP constructs a reasoning tree guided by a world model to identify high-reward reasoning paths that properly balance exploration and exploitation. 
In various challenging reasoning tasks, 
RAP combined with LLaMA-33B even outperformed chain-of-thought (CoT) prompting with GPT-4, achieving a 33\% relative improvement in the plan generation task \cite{hao2023reasoning}.

\subsection{World Model and Next-generation Edge Intelligence}

In edge intelligence deployments such as autonomous drones, smart‑factory robots, and mixed‑reality headsets, devices face sparse on‑device observations, task‑specific retraining overhead, and intermittent or low‑rate connectivity. 
Deploying a world model at the edge mitigates these constraints by completing occluded regions or forecasting future frames to sustain robust perception and prediction from minimal raw input.
Techniques in 6G communication, such as semantic compression, which transmits only task‑relevant bits by packetizing latent codes and imagined trajectories, shrink spectrum usage without degrading decision quality, while edge–cloud split inference leveraging holographic multi-input multi-output (MIMO) and sub‑THz backhaul lets heavy layers of the world model execute in micro‑datacenters and latency‑critical layers remain on‑device.
In short, world models confer foresight, causal reasoning, and sample-efficient learning on edge devices, while 6G supplies the bandwidth, low latency, and semantic protocols to train, distribute, and maintain those models at scale.

\section{World Model for Wireless Edge Intelligence Optimization}
\label{sec:frame}


\subsection{Spatio-temporal Optimization Challenges in Wireless Intelligence Networks}
\label{sec:challenge}

Spatio-temporal optimization is an important issue in wireless intelligent networks, especially in advanced LAWNs, such as UAV-based communication systems, which face unique challenges due to their dynamic, time-varying nature. For example, in LAWNs, the communication links fluctuate with unpredictable environmental conditions, interference, and UAV mobility, cascading effects on network performance in the future. 
In summary, the key challenges in wireless edge intelligence networks include:
\begin{itemize}
    \item \textbf{Highly Dynamic Environments:} 
    LAWNs experience rapid changes in topology and environmental conditions. UAVs functioning as flying base stations must dynamically adapt to maintain coverage and throughput. 
    
    \item \textbf{Sequential Decision Coupling:} Actions are coupled across time, influencing the state of the network in subsequent moments. Optimizing over time thus requires foresight, anticipating the environment changing.
    


    \item \textbf{Partial Observability and Uncertainty:} In practice, the agent may not know future user demand or channel conditions with certainty. It must infer and learn spatio-temporal patterns 
    to make proactive decisions. 

\end{itemize}

These challenges highlight the need for an intelligent optimization framework that plans ahead and adapts to the spatio-temporal dynamics of wireless intelligence networks. 
Traditional RL algorithms such as Deep Q-Network (DQN) are model-free, which 
learn policies by direct trial-and-error in the environment, without learning an internal model \cite{li2019board}.
In a complex wireless network, a DQN agent would need to experience a large variety of network conditions and long sequences of actions to eventually stumble upon an effective long-term strategy. 
This is impractical in real-world scenarios where each trial can degrade service or incur cost. 
Moreover, model-free agents lack an explicit mechanism for spatio-temporal reasoning, whose decisions are reactive based on the current state.
Incorporating a world model addresses these limitations by giving the agent the ability to simulate and evaluate imagination scenarios internally. Instead of relying solely on incremental learning from real experiments, the agent can predict the outcome of potential action sequences in its learned simulator of the network.
This provides several key benefits for spatio-temporal optimization:
\begin{itemize}
    \item \textbf{Multi-step Planning:} The agent can generate imagined future trajectories 
    and assess the long-term consequences of actions. By planning over these latent trajectories, the agent develops far-sighted behaviors that account for future rewards, rather than just immediate effects. 
    

    \item \textbf{Sample Efficiency:} 
    With a world model, the agent learns the dynamics from real data and then reuses that knowledge by simulating numerous virtual trajectories. This drastically cuts down on the interactions with environments, especially when experiments on physical UAV networks are costly or time-limited.


    \item \textbf{Handling Uncertainty and Partial Observability:} The world model’s latent state can serve as a form of memory, integrating information over time. Agents can then infer unobserved aspects of the system, such as the underlying channel condition or user mobility patterns. 
    
    
    
    
\end{itemize}

In summary, the world model provides the foresight and structured memory that align naturally with the spatio-temporal optimization needs of advanced wireless networks.

\subsection{Proposed Framework}

\begin{figure*}[htp]
    \centering
    \includegraphics[width= 0.85\linewidth]{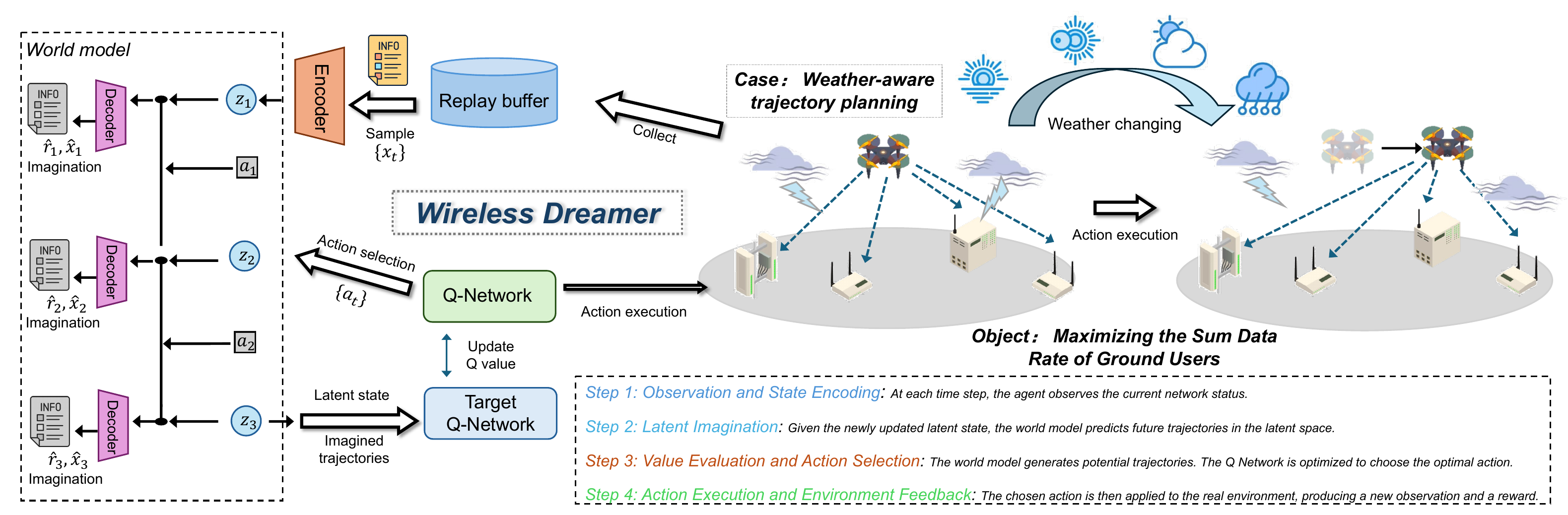}
    \caption{The workflow of the proposed framework. The left part is the model structure of Wireless Dreamer, including a world model, Q-network, and target Q-network. The bottom part presents the continuous processes of Wireless Dreamer. The right part illustrates the weather-aware trajectory planning in a UAV-assisted scenario in LAWNs.}
    \label{fig:frame}
\end{figure*}

To realize these benefits, we introduce Wireless Dreamer, a Q-learning framework integrating a world model. 
It is a framework tailored to solve spatio-temporal optimization problems in networks of UAVs and other low-altitude wireless platforms with a discrete action space. 
As shown in Fig. \ref{fig:frame}, our framework consists of three main components analogous:
\begin{itemize}
    \item \textbf{Latent World Model:} a learned dynamics model that predicts how the wireless network state changes over time. 
    This model ingests the current observation, such as channel measurements, UAV locations, queue lengths, and the previous latent state, and produces a predicted next state and reward. 
    

    \item \textbf{Q Network:} a neural network that estimates the long-term value of a given latent state.
    In Wireless Dreamer, the Q Network is trained on simulated trajectories from the world model, learning to assess network states that may never have been directly observed. 

    \item \textbf{Target Q Network:} a periodically updated replica of the primary Q Network, used to generate stable bootstrap targets during value learning.

\end{itemize}

\subsection{Wireless Dreamer Workflow Overview}

To illustrate how the components come together, Fig. \ref{fig:frame} and the following steps outline the Wireless Dreamer workflow\footnote{For more detailed information, including source codes and experiment settings, please refer to this tutorial page: {https://changyuanzhao.github.io/Wireless\_Dreamer}}. 
The process continually cycles through observation, latent model update, imagination/planning, action selection, and environment feedback, as detailed below:
\begin{enumerate}
    \item \textbf{Observation and State Encoding:} At each time step $t$, the agent observes the current network status $x_t$. The observation is fed through an encoder to update the agent’s latent state $z_t$.
    

    \item \textbf{Latent Imagination:} Given the latent state $z_t$, the world model predicts future trajectories in the latent space by its world model.
    Conditioned on the selected action $a_t$, it generates a sequence of latent states and intermediate rewards, which can be decoded into reconstructed observations $\hat{x}_t$ and predicted rewards $\hat{r}_t$, respectively.



    \item \textbf{Value Estimation and Action Selection:} 
    The world model generates potential latent states and corresponding predicted rewards. The Q Network assesses each candidate by calculating \( Q(z_{t+1}, a_t) \) for discrete actions.
    
    
   

    \item  \textbf{Action Execution and Environment Feedback:} The agent then picks action $a_t$ based on \( Q(z_{t+1}, a_t) \) to execute in the real environment, producing a new observation $x_{t+1}$ and a reward $r_t$.
    Over time, 
    the world model becomes more accurate in its predictions, and the policy becomes more adept at choosing optimal actions. 
    
\end{enumerate}

Throughout this workflow, the Wireless Dreamer framework effectively integrates model-based planning with RL. By continuously learning and planning in the latent space, it navigates the spatio-temporal complexity of wireless network optimization. The agent can anticipate network behavior and plan multi-step strategies, addressing the challenges identified in Section \ref{sec:challenge}.


\section{Case Study: Weather-aware UAV Trajectory Planning}


In this section, we present a case study on weather-aware UAV trajectory planning and demonstrate how the proposed framework enhances spatio-temporal optimization.

\subsection{System Model}

We consider a UAV-assisted wireless communication scenario in LAWNs, where a single UAV acts as a mobile base station to provide downlink coverage for multiple ground users under dynamic and spatially varying weather conditions, as shown in Fig. \ref{fig:frame}. The UAV traverses a two-dimensional area over a fixed time horizon and dynamically adjusts its position to optimize communication performance while responding to evolving environmental factors such as wind intensity.

This scenario involves one UAV and $N$ users randomly distributed across a $64 \times 64$ grid-based area, where each grid is $4~m\times 4~m$. The UAV operates over a 28 GHz mmWave band with a total bandwidth of 100 MHz and a transmission power of 30 dBm. 
The weather is modeled by the Gaussian Field Model, which simulates a drifting Gaussian hotspot to represent dynamic environmental factors such as wind speed \cite{chen2013wind}. The hotspot's position evolves over time, directly affecting the communication path loss.
The UAV aims to maximize the total downlink capacity across users within $T$ time steps, while its trajectory is influenced by the need to avoid adverse weather zones and turbulence.
This case study considers 10 users and a time horizon of 100 steps.

\subsection{Numerical Results}

\subsubsection{Experimental Setup}

Our experiment is
conducted on a Linux server equipped with an Ubuntu 22.04 operating system and powered 
an NVIDIA RTX A6000 GPU. 
We benchmark the proposed method with DQN \cite{li2019board} and a random policy.

\subsubsection{Performance Analysis}

\begin{figure}[htbp]
\centering
\includegraphics[width=0.62\linewidth]{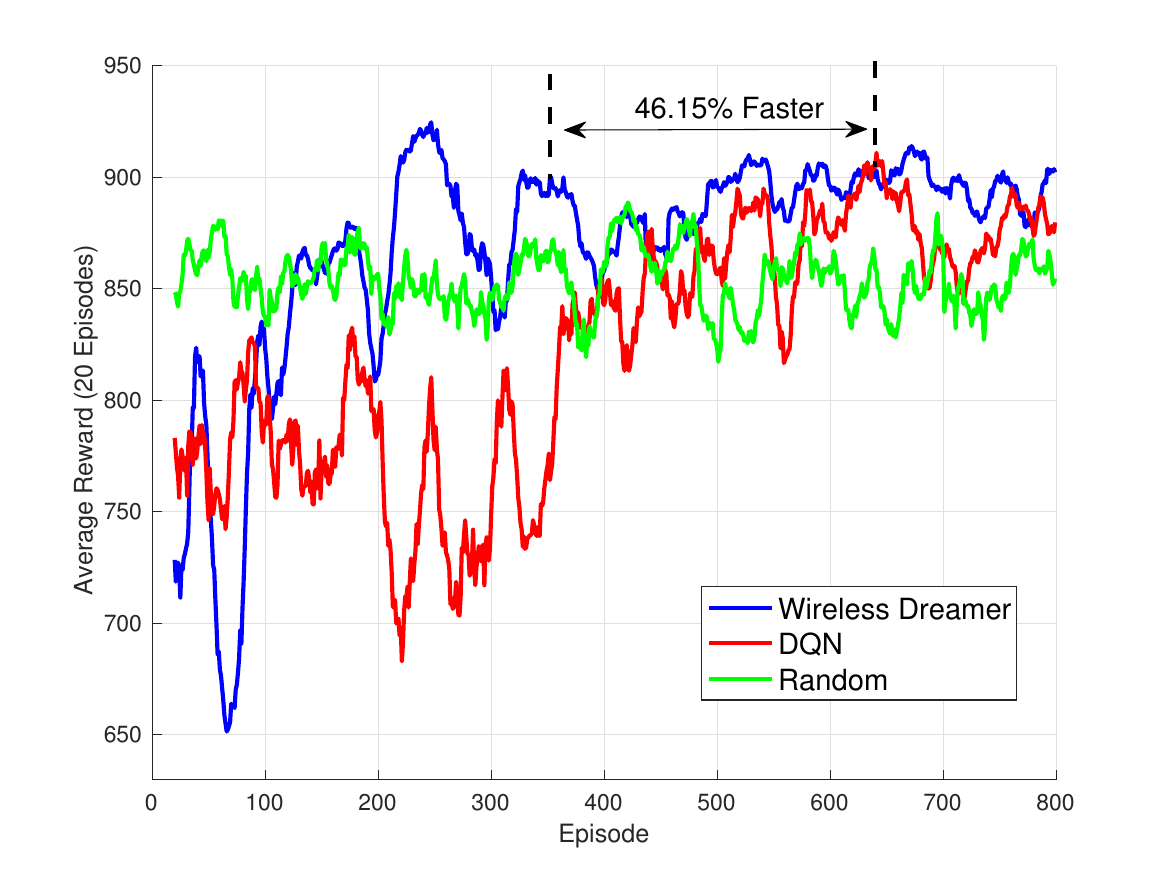} 
  \caption{Comparison of average episodic rewards}
\label{fig:ex1}
\end{figure}

\begin{figure}[htbp]
\centering
\includegraphics[width=0.62\linewidth]{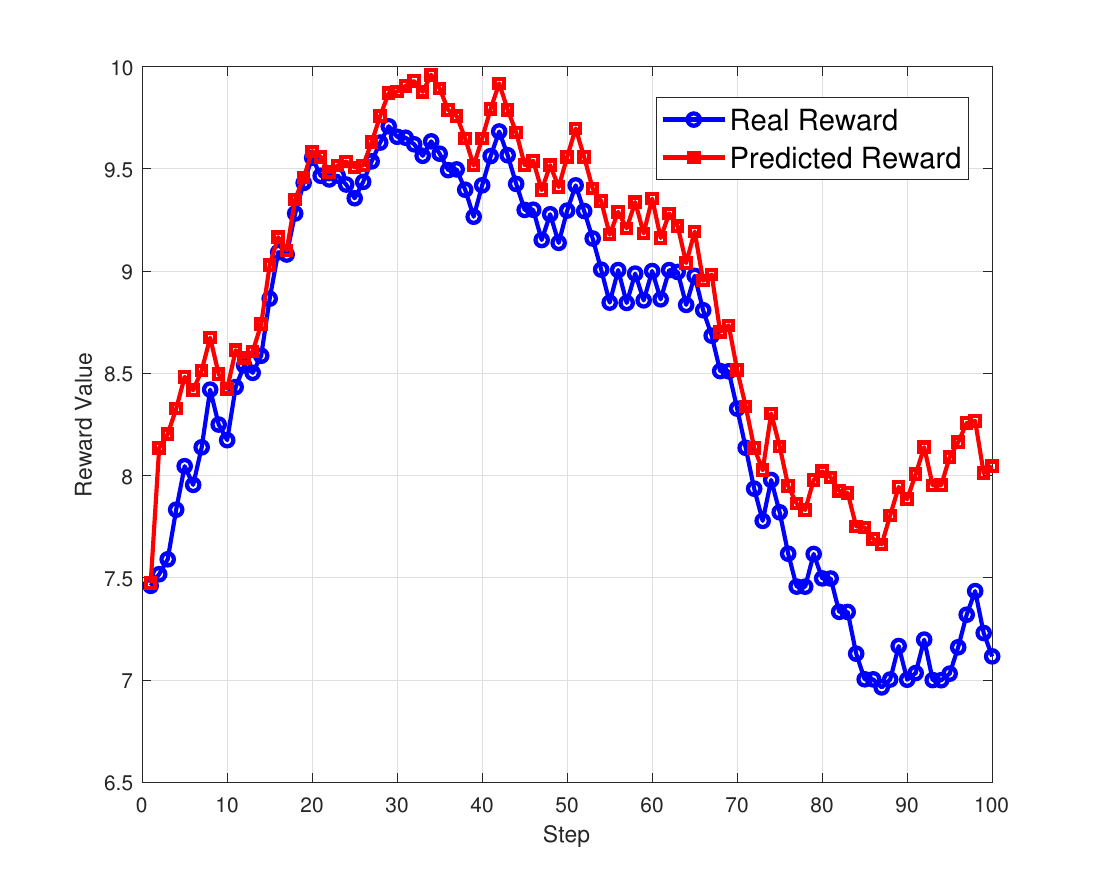} 
  \caption{Comparison between real and predicted rewards}
\label{fig:ex2}
\end{figure}

The average episodic rewards of the three methods are shown in Fig. \ref{fig:ex1}, where each point represents the average reward over 20 consecutive episodes.
The results demonstrate that our proposed algorithm, Wireless Dreamer (blue curve), rapidly discovers a superior policy and converges to a stable performance. By episode 250, its average reward reaches 923.55, clearly surpassing the optimal average reward of 829.04 attained by DQN (red curve) during the same training episodes. Furthermore, Wireless Dreamer's coverage by approximately episode 350 is about 46.15\% faster than DQN, which does not level off until roughly episode 650.
This performance gap stems from Wireless Dreamer’s capacity to employ its learned world model for trajectory imagination and prediction, extracting far richer trajectory information than that available through direct environment interactions alone. 
This benefit is critical for scenarios with restricted sensing opportunities, such as the weather-aware UAV communication environment examined in this work, where exhaustive collection of environmental information is seldom feasible.

Moreover, we compare the real rewards and the predicted rewards of our proposed method in the evaluation stage.
As illustrated in Fig. \ref{fig:ex2}, the predicted rewards track the real returns with high fidelity during evaluation. Over the entire test set, the mean absolute error is
0.359 ± 0.262
, and the maximum deviation observed is 1.059, yielding an average relative error of roughly 5\%. Remarkably, during the first 30 decision steps, the predictions are nearly equal to the real rewards, underscoring the world model’s ability to capture agent state and reward dynamics.





\subsubsection{Limitations}

As discussed in Section \ref{sec:frame}, the defining advantage of Wireless Dreamer over conventional RL methods is that real interactions are used to fit the world model, after which the agent learns by generated trajectories.
However,
any policy learned from the model inevitably inherits its prediction errors, which compound over long-horizon prediction and can degrade performance, as shown in Fig. \ref{fig:ex2}. When abundant real data are available, 
a model-free controller that learns directly from the environment (i.e., DQN) may outperform Wireless Dreamer because it is not subject to modelling bias.
Therefore, in future work, we will focus on reducing modelling error and expanding the world model to applications with limited interaction with the environment.


\section{Conclusion}

In this paper, we have explored the concept, architecture, and deployment of world models as a foundation for cognitive agents in future edge intelligence systems. We have provided a detailed taxonomy and comparison with adjacent paradigms such as digital twins, the metaverse, and foundation models, positioning world models as internal cognitive mechanisms for agents.
We have introduced Wireless Dreamer, a world model-enhanced Q-learning framework that significantly improves sample efficiency and temporal foresight in wireless edge environments. Using a case study of weather-aware UAV trajectory planning, we have shown that Wireless Dreamer accelerates convergence by 46.15\% compared to traditional DQN and achieves high fidelity in reward prediction.
In summary, world models offer a promising path toward building cognitive systems in the future network edge.

\bibliography{Ref}

\end{document}